\PassOptionsToPackage{table,dvipsnames}{xcolor}
\documentclass[runningheads]{llncs}
 
\usepackage{eccv}

\usepackage{eccvabbrv}

\usepackage{graphicx}
\usepackage{booktabs}

\usepackage[accsupp]{axessibility}  
\usepackage{multirow}
\usepackage{array}
\usepackage{booktabs}
\usepackage{makecell}
\usepackage{bbding}
\usepackage{diagbox}
\usepackage{soul}
\usepackage{wrapfig}
\usepackage[ruled,linesnumbered]{algorithm2e} 
\usepackage{hyperref}

\usepackage{hyperref}

\usepackage{orcidlink}

\begin{document}
\setlength{\floatsep}{3pt plus 1pt minus 2pt}
\setlength{\textfloatsep}{3pt plus 1pt minus 2pt}
\setlength{\intextsep}{3pt plus 1pt minus 2pt}
\title{A$^{3}$lign-DFER: Pioneering Comprehensive Dynamic Affective Alignment for Dynamic Facial Expression Recognition with CLIP} 

\titlerunning{A$^{3}$lign-DFER}

\author{Zeng Tao\inst{1} \and
Yan Wang\inst{1}\Envelope \and
Junxiong Lin\inst{1} \and
Haoran Wang\inst{1} \and
Xinji Mai\inst{1} \and
Jiawen Yu\inst{1} \and
Xuan Tong\inst{1} \and
Ziheng Zhou\inst{2} \and
Shaoqi Yan\inst{1} \and
Qing Zhao\inst{1} \and
Liyuan Han\inst{3} \and
Wenqiang Zhang\inst{4,5}\Envelope}

\authorrunning{Z.~Tao et al.}

\institute{Shanghai Engineering Research Center of AI \& Robotics, Academy for Engineering \& Technology, Fudan University, Shanghai, China. \and School of Information Science and Technology, Fudan University, Shanghai, China. \and
Institute of Automation, Chinese Academy of Sciences, Beijing, China. \and
Engineering Research Center of AI \& Robotics, Ministry of Education, Academy for Engineering \& Technology, Fudan University, Shanghai, China. \and Shanghai Key Lab of Intelligent Information Processing, School of Computer Science, Fudan University, Shanghai, China\\ \Envelope Corresponding Authors}

\maketitle
\begin{abstract}
  The performance of CLIP in dynamic facial expression recognition (DFER) task doesn't yield exceptional results as observed in other CLIP-based classification tasks. While CLIP's primary objective is to achieve alignment between images and text in the feature space, DFER poses challenges due to the abstract nature of text and the dynamic nature of video, making label representation limited and perfect alignment difficult. To address this issue, we have designed A$^{3}$lign-DFER, which introduces a new DFER labeling paradigm to comprehensively achieve alignment, thus enhancing CLIP's suitability for the DFER task. Specifically, our A$^{3}$lign-DFER method is designed with multiple modules that work together to obtain the most suitable expanded-dimensional embeddings for classification and to achieve alignment in three key aspects: affective, dynamic, and bidirectional. We replace the input label text with a learnable Multi-Dimensional Alignment Token (MAT), enabling alignment of text to facial expression video samples in both affective and dynamic dimensions. After CLIP feature extraction, we introduce the Joint Dynamic Alignment Synchronizer (JAS), further facilitating synchronization and alignment in the temporal dimension. Additionally, we implement a Bidirectional Alignment Training Paradigm (BAP) to ensure gradual and steady training of parameters for both modalities. Our insightful and concise A$^{3}$lign-DFER method achieves state-of-the-art results on multiple DFER datasets, including DFEW, FERV39k, and MAFW. Extensive ablation experiments and visualization studies demonstrate the effectiveness of A$^{3}$lign-DFER. The code will be available in the future.
\end{abstract}

\section{Introduction}
\label{sec:intro}


With the advancement of large-scale models utilizing contrastive learning, such as CLIP~\cite{radford2021learning}, there has been a notable enhancement in image classification tasks~\cite{lin2023multimodality,Huang_2023_ICCV,Abdelfattah_2023_ICCV}. This improvement stems from CLIP's ability to harness a substantial amount of prior knowledge for aligning images (samples) with texts (prompts and labels) within the feature space. However, this approach is not entirely effective for Dynamic Facial Expression Recognition (DFER) tasks, failing to achieve the same level of excellence as in other applications. DFER, significant in areas like human-computer interaction and health monitoring, faces challenges in alignment within the CLIP framework due to its complex nature. In essence, CLIP aligns semantically similar texts and images in the feature space through contrastive learning, paving the way for adaptability across various downstream tasks. However, aligning textual labels and facial expression videos for DFER tasks in the feature space using a pre-trained CLIP model is not straightforward.

\begin{figure}[tb]
  \centering
  \includegraphics[height=4.7cm]{ 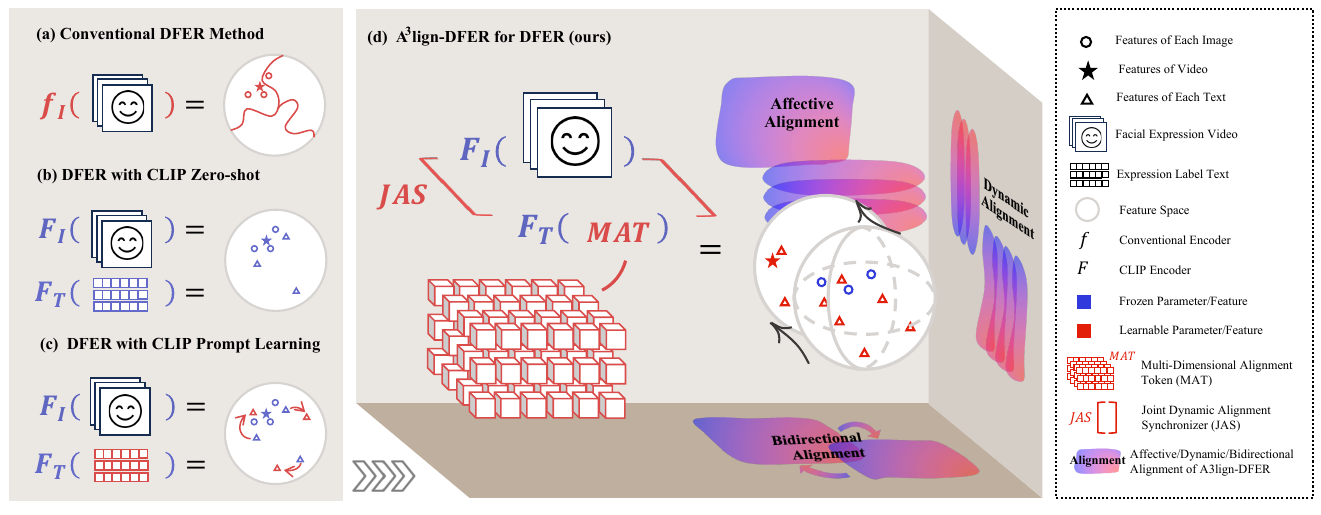}
    \caption{The comparison between the basic paradigm of dynamic facial expression recognition and our A$^{3}$lign-DFER. (a) represents traditional DFER methods~\cite{wang2023rethinking, wang2022dpcnet, li2023intensity}, training discriminative models to obtain classification results. (b) represents the CLIP method~\cite{radford2021learning}, classifying by comparing the similarity in the feature space between videos and label text encoding. (c) represents the CLIP-based prompt learning method~\cite{zhou2022learning, zhou2022conditional, li2023cliper, zhao2023prompting}, using learnable embeddings instead of fixed label embeddings. (d) represents our proposed A$^{3}$lign-DFER, achieving the best match between expression videos and classes through expanded-dimensional learning and comprehensive dynamic affective alignment.}
    \label{fig:onecol}
\end{figure}


The alignment challenges posed by the complex characteristics of DFER manifest in several ways. Firstly, facial expressions represent an abstract concept, with labels acting as abstract proxies for classes rather than detailed expressions of feature distribution. Thus, aligning simplistic, abstract facial expression label texts with diverse, intricate facial expression images in the feature space fails to yield an affective alignment. Secondly, facial expression videos contain dynamic elements, with various changes often reflecting essential aspects of the expressions. Addressing how to represent these dynamic shifts in the feature space and align labels and samples along the temporal dimension remains a critical, yet under-explored, area. Therefore, developing affective and dynamic mappings between facial expression videos and expression classes in both videos and texts is crucial for maximizing the use of large model's prior information and optimizing DFER model performance.

In light of these factors, we have developed the A$^{3}$lign-DFER method, as shown in Fig.~\ref{fig:onecol}. Differing from previous methods in (a)(b)(c), this approach in (d) aims to obtain the most suitable expanded-dimensional embeddings for DFER and achieve comprehensive alignment between expression videos and class label texts within the pre-trained CLIP model. Specifically, our approach includes designs that focus on Affective, Dynamic, and Bidirectional Alignment. Firstly, we introduced a novel concept, the learnable Multi-dimensional Alignment Tokens (MAT), to supersede traditional class labels and prompts, adding a third dimension to the alignment process. These tokens autonomously seek the most relevant embeddings for expression classes within the feature space, moving beyond limited abstract texts to achieve optimal affective alignment. Additionally, the expanded dimension's innovative design not only offers a wider range of parameter possibilities but also establishes a foundation for subsequent dynamic alignment. Secondly, after processing through the frozen CLIP text and image encoders, we devised a Joint Dynamic Alignment Synchronizer (JAS), enabling the synchronization of text and image features. This synchronizer captures the rich expressive information within the temporal variations of the videos, integrating it into the class features represented by MAT. The unique combination of MAT and JAS in our model facilitates Affective and Dynamic Alignment of expression videos and label texts under the CLIP classification paradigm. However, direct training of both text and image data flows at the same time, each with its learnable parameters, does not yield optimal alignment results. Consequently, we have designed a Bidirectional Alignment Training Paradigm (BAP), ensuring gradual and steady training of both flows' parameters, thereby achieving effective alignment in the feature space. Through A$^{3}$lign-DFER, we obtain the most suitable expanded-dimensional embeddings (MAT) for DFER and achieve a tripartite alignment—affective, dynamic, and bidirectional—between expression videos and abstract learnable labels, leading to enhanced DFER outcomes. 


In summary, our main contributions are three-fold:
\begin{itemize}
    \item We propose a comprehensive alignment paradigm for DFER (namely A$^{3}$lign-DFER), which can explore the alignment of expression videos and abstract labels in the CLIP paradigm from the perspectives of affective, dynamic, and bidirectional aspects.
    \item We introduce a novel labeling scheme and a tailored training paradigm. Through MAT and JAS, an expended-dimensional embedding was trained as a textual label, using content that is diverse and contains dynamic information to correct and enrich the label's content. Furthermore, we have developed a training paradigm BAP for the DFER method, ensuring gradual and steady training of parameters in both text and image data flows.
    \item We conduct extensive experiments on three in-the-wild DFER datasets ( DFEW, FERV39k, and MAFW) to prove the effectiveness and the superiority of the proposed A$^{3}$lign-DFER, which achieves state-of-the-art results compared with other methods.
\end{itemize}



\vspace{-0.2cm}
\section{Related Work}
\label{sec:relatedwork}
\vspace{-0.1cm}
\subsection{Dynamic Facial Expression Recognition}

DFER task is challenging, due to the requirements of dealing with the abstraction of facial expressions and the dynamics of videos. Some conventional methods use 3D CNNs~\cite{cohen2003facial, yu2018spatio} to extract spatiotemporal features from sequences, while others employ cascading CNN-LSTM~\cite{fan2016video, liu2020saanet, vielzeuf2017temporal, yu2020facial} and CNN-transformer~\cite{li2022nr, li2021jdman, tao2023freq} structures. Liu et al.~\cite{liu2022clip} introduced a twin CNN-LSTM cascading structure with action unit attention for encoding paired videos, allowing the fine differentiation of different expressions. Wang et al.~\cite{wang2022dpcnet} proposed DPCNet, a method for learning appearance features and dynamic temporal information within cascading structures, delving deep into the exploration and learning of latent representations selected from four segments of a video.

Starting from the data, these methods train specialized classification models focusing on spatial and temporal feature extraction, aiming to map inputs to outputs. As large models develop, vast pretraining data becomes prior knowledge for classification tasks. Utilizing this prior information in dynamic facial expression recognition is crucial. This study focuses on applying the contrastive learning pre-trained model CLIP in DFER, aiming for comprehensive dynamic affective alignment using CLIP.

\vspace{-0.1cm}
\subsection{CLIP in Classification Tasks}

The Contrastive Language-Image Pre-Training (CLIP)~\cite{radford2021learning} relies on a vast corpus of image-text pairs for its foundation in contrastive learning, resulting in robust pre-trained image and text encoders that exhibit remarkable feature extraction capabilities~\cite{chen2023disco}. These encoders map images and text to close feature space positions, useful in applications~\cite{chen2023clip2scene, he2023clip, sanghi2023clip, tao2023galip, tschannen2023clippo} like semantic segmentation~\cite{liang2023open, lin2023clip}, visual retrieval~\cite{pei2023clipping, sain2023clip}, and 3D learning~\cite{huang2023shapeclipper, hyung2023local}. In image classification~\cite{conde2021clip, zhang2022tip, ali2023clip}, CLIP combines class labels with a shared prompt for text encoding, determining classification by feature similarity. Extensions like CoOp~\cite{zhou2022learning} and CoCoOp~\cite{zhou2022conditional} make text prompts learnable for optimal embeddings. In video processing~\cite{luo2022clip4clip, lin2022frozen}, CLIP adapts with temporal processing and fine-tuning~\cite{udandarao2023sus, lin2023multimodality, Wei_2023_ICCV}. Nevertheless, it's important to note that these approaches entail the fine-tuning~\cite{Liu_2023_CVPR} of the CLIP model itself, which could potentially be detrimental to the model's integrity.

Addressing dynamic facial expressions, current datasets are limited in scale and plagued by label noise, making simple fine-tuning of the CLIP model impractical. This paper introduces a novel strategy: keeping CLIP's pre-trained image and text encoders frozen and reverse-fitting the large model using a data module. This method aligns with CLIP's structure, utilizes its prior knowledge, and aims to enhance performance in DFER tasks.

\vspace{-0.1cm}
\subsection{Prompt Learning and Answer Learning}

Prompt learning is a method that has emerged in response to the development of Large Language Models (LLM)~\cite{huang2023vop, yao2023visual, guo2023zero, liu2023hierarchical}. In contrast to traditional fixed prompts, prompt learning treats them as learnable parameters with the goal of finding the most suitable embeddings for the model to employ~\cite{zhou2022learning, zhou2022conditional}. These embeddings may direct the model toward language features that are completely incomprehensible to humans. In addition to automatically learning prompts, some research has also attempted to learn answers. One of what set this work apart is that it parametrizes all text embeddings, obtaining the embeddings that best match the classification task.

Previous prompt learning methods, mainly tailored for image-based tasks, lack attention to video-based tasks, especially embedding dimensions. Our approach innovatively adopts multidimensional embeddings to optimize dynamic information use and achieve full-chain alignment.

\vspace{-0.1cm}
\section{Method}
\label{sec:method}
\vspace{-0.1cm}
\subsection{Overview}

In A$^{3}$lign-DFER, we have specifically designed learnable Multi-dimensional Alignment Tokens (MAT) and a Joint dynamic Alignment Synchronizer (JAS), along with a tailored Bidirectional Alignment training Paradigm (BAP), as shown in Fig.~\ref{fig:method}. For an input facial expression video $V$, its individual $F$ frames are separately fed into a frozen CLIP image encoder, resulting in $F$ image features. For MAT, each of its $Cls$ classes and groups of $Snt$ sentences per class is input into a frozen CLIP text encoder, yielding $Cls\times Snt$ text features. The video feature set containing $F$ images and the $Cls$ groups of $Snt$ text features are then fed into JAS, resulting in semantic and dynamic features for each video and each group of class text. The similarity in feature space is computed between these video features and each group of class text features obtained by MAT and JAS, with the highest similarity determining the classification class. The training paradigm within the method follows BAP.
\vspace{-0.1cm}
\subsection{Multi-Dimensional Alignment Token}

Using CLIP for zero-shot video classification involves inputting each frame into the image encoder to obtain features and averaging them for video features. Different label texts are encoded for each class. When fine-tuning CLIP for downstream tasks, it usually requires high-quality and abundant data. However, for DFER, using small or low-quality datasets for fine-tuning large models can lead to poor results due to dataset size and label noise. Facial expressions are abstract, and single-label texts may not capture the richness of acoustic information in expression classes.

To make the most of CLIP's prior knowledge and powerful feature extraction capabilities while adapting to DFER tasks, inspired by the work of CoOp~\cite{zhou2022learning}, CoCoOp~\cite{zhou2022conditional} and CLIPER~\cite{li2023cliper}, we adopt a prompt learning method to learn text embeddings suitable for dynamic facial expression classification and pre-trained CLIP. It is important to highlight that, in the context of the DFER task, our method uniquely configures all text tokens as learnable. This approach aims for a broad and dynamically adaptable framework across various dimensions, representing a novel aspect that previous studies have not explored.

\begin{figure}[tb]
  \centering
  \includegraphics[height=6.3cm]{ 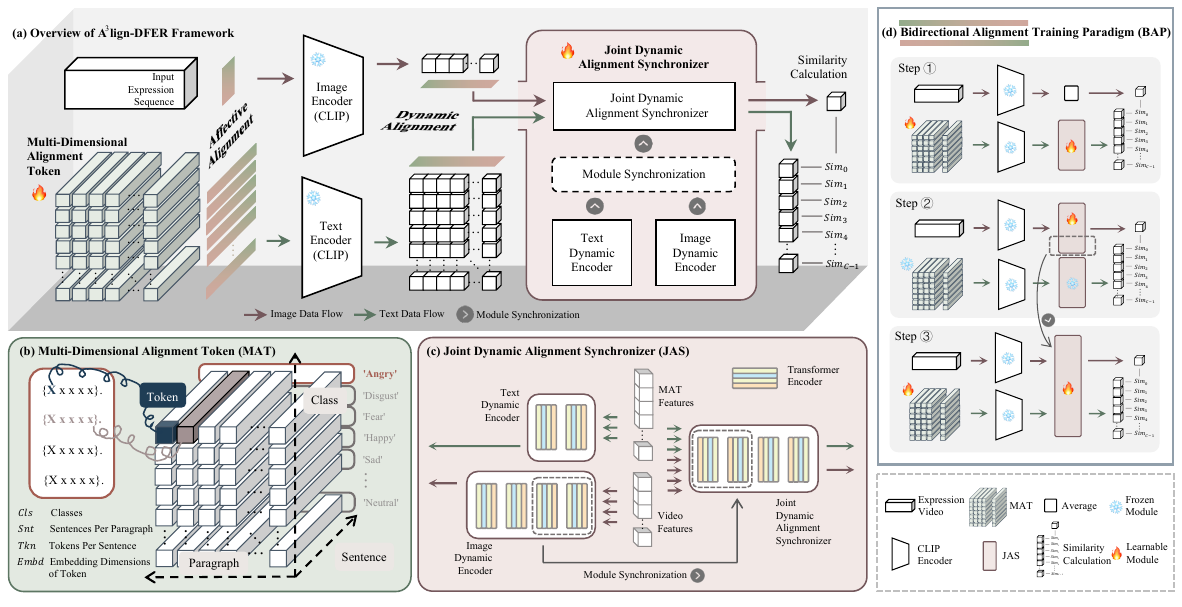}
  \caption{A$^{3}$lign-DFER Method Workflow Overview. (a) Illustrates the A$^{3}$lign-DFER process, processing facial expressions and MAT through CLIP encoders, then into JAS for class-specific video features and classification. (b) Details MAT's structure and dimension meanings. (c) Shows JAS module construction. (d) Highlights the Bidirectional Alignment Training Paradigm (BAP) in our method.
  }
  \label{fig:method}
\end{figure}

Specifically, as shown in Fig.~\ref{fig:method} (b), we set the entire text embeddings input into the CLIP Text Encoder $E^{\mathrm{CLIP}}_{\mathrm{Text}}$ as learnable parameters. We update these parameters through a contrastive learning loss by backpropagation to find the embeddings in the embedding space that best match the class. During inference, the trained embeddings are input into $E^{\mathrm{CLIP}}_{\mathrm{Text}}$ to obtain encoded features.

To align with the subsequent extraction of video temporal information, we have designed the text embedding part in a specific way. In addition to arranging multiple embeddings in the sentence dimension, we extend them in the third dimension, aiming to capture the dynamic information inherent in changing facial expressions. Specifically, for each embedding element $f_{ijkl}$, it is demonstrated as follows:

\begin{equation}\label{eq1}
  \begin{aligned}
    f_{ijkl}  = MAT&[c_{i}, s_{j}, t_{k}, e_{l}],\\
    d\in \left \{ 0, 1, \dots, D-1 \right \} , d=&i, j, k, l,D= Cls,Snt,Tkn,Embd,
  \end{aligned}
\end{equation}
where $MAT$ represents the multi-dimensional learnable tokens, $Cls$, $Snt$, $Tkn$, $Embd$ represents the numbers of classes, sentences in one class, Tokens in one sentence, and embedding elements in one token respectively. The CLIP encoding process of the $i^{th}$ class and $j^{th}$ sentence in multi-dimensional learnable tokens $Feat_{ij}^{text}$ is demonstrated as follows:

\begin{equation}\label{eq2}
  \begin{aligned}
    Feat^{\mathrm{text}}_{ij} = E^{\mathrm{CLIP}}_{\mathrm{Text}}(\mathrm{concat}(MAT[c_{i}, s_{j}, :Tkn, :Embd], \\
    \mathrm{zeros}[Tkn_{\mathrm{max}}-Tkn, Embd]))
  \end{aligned}
\end{equation}
where $E^{\mathrm{CLIP}}_{\mathrm{Text}}$ represents the pre-trained CLIP text encoder, $Tkn_{\mathrm{max}}$ represents for the max context length for Transformer, which is set to 77 in CLIP. In reality, $Tkn_{\mathrm{max}}$ is 74, accounting for the removal of the $<$SOS$>$, $<$EOS$>$, and the period.

Empirical evidence demonstrates that in A$^{3}$lign-DFER, increasing the learnable token dimensions yields better results than extending parameters in a single dimension, which is demonstrated in Experiment Section~\ref{sec:experiment}. By elevating the dimensionality, these high-dimensional learnable tokens gain the ability to uncover the most optimal embeddings for expression video classification in both semantic and dynamic aspects, leading to the groundwork for subsequent feature extraction. On the other hand, as a pre-trained model, the CLIP pretraining text encoder's limitation stems from its fixed max context length, which is demonstrated in Eq.~\ref{eq2} as $Tkn_{\mathrm{max}}$, of 74, rendering it impossible to enhance the performance of the downstream module by extending the number of tokens in the context dimension. Elevating the dimensions provides a workaround for this constraint, opening up additional possibilities for designing the downstream module.

\vspace{-0.1cm}
\subsection{Joint Dynamic Alignment Synchronizer}

Previous attempts to incorporate dynamic feature extraction modules into various stages of CLIP, such as adding modules after CLIP or embedding them within CLIP, have not yielded outstanding results. They also have significant limitations and do not integrate well with text.

Building upon the foundation of the previous design of MAT, we have designed the dynamic encoder part accordingly. 
In fact, by aligning the video with MAT in the temporal dimension, we can train MAT to implicitly learn dynamic-related features. Using JAS, we can force MAT to learn spatially consistent, temporally aligned facial expression class information across multiple dimensions.

\textbf{Dynamic Alignment Synchronizing.} Specifically, in the temporal dimension, we input images and MAT into the CLIP encoders separately to obtain different temporal features. The image encoding process is demonstrated as follows:

\begin{equation}\label{eq3}
  \begin{aligned}
    Feat^{\mathrm{image}}_{j'} = E^{\mathrm{CLIP}}_{\mathrm{Image}}(V[F_{j'}, :Channel, :H, :W]),
  \end{aligned}
\end{equation}
where $V$ represents the expression video, $E^{\mathrm{CLIP}}_{\mathrm{Image}}$ represents the pre-trained CLIP image encoder, $F_{j'}$ represents the $j'^{th}$ frame in the video. These features are then processed through implicit dynamic encoders $DE^{\{\mathrm{Text}, \mathrm{Video}\}}$ to obtain temporal encoded features as follows:

\begin{equation}\label{eq4}
  \begin{aligned}
    Feat^{\mathrm{cls}}_{i} = DE^{\mathrm{Text}}(Feat^{\mathrm{text}}_{ij} \ for \ j \ in \ (0, S)), \\
     Feat^{\mathrm{video}} = DE^{\mathrm{Video}}(Feat^{\mathrm{image}}_{j'} \ for \ j' \ in \ (0, F))
  \end{aligned}
\end{equation}
The similarity $Sim_{i}$ of expression video and $i^{th}$ class is calculated between the features obtained by encoding image and text features in the temporal dimension as follows:

\begin{equation}\label{eq5}
  \begin{aligned}
    Sim_{i} = \cos (\theta) = \frac{Feat^{\mathrm{cls}}_{i}\cdot Feat^{\mathrm{video}}}{\left \| Feat^{\mathrm{cls}}_{i} \right \| \left \|  Feat^{\mathrm{video}}\right \| }
  \end{aligned}
\end{equation}
The prediction probability $cls_{i}$is then:

\begin{equation}\label{eq6}
  \begin{aligned}
    p(cls_{i}|V)=\frac{\exp (sim_{i}/\tau )}{ {\textstyle \sum_{i=0}^{Cls-1}} \exp (sim_{i}/\tau )} 
  \end{aligned}
\end{equation}
$\tau$ is temperature parameter. Cross-entropy loss is used as the learning objective.


\textbf{Construction of JAS.} In the A$^{3}$lign-DFER, we adopt the widely recognized transformer as the fundamental framework for JAS, as shown in Fig.~\ref{fig:method} (c). In the initial phases, we conduct separate training for the text and image branches. When we reach the fusion stage, our approach involves using the first two transformer layers of the well-established image dynamic encoder as a base and then proceeding to add layers step by step and retrain MAT. This decision stems from the fact that expression videos, used as the dataset input, contain more extensive semantic information. For further training and inference details, please refer to Section~\ref{sec:BAP} and~\ref{sec:Inference}.

\vspace{-0.1cm}
\subsection{Bidirectional Alignment Training Paradigm}
\label{sec:BAP}

However, for CLIP-based methods, direct backpropagation can lead to training stagnation when both labels and samples contain a large number of trainable parameters. Therefore, we have designed a bidirectional alignment training paradigm that aims for stable parameter positioning, from overall to local search, as shown in Fig.~\ref{fig:method}(d).


The specific steps are as follows: In the first step, we freeze the CLIP image and text encoders. We input each frame of images from the video and embeddings in each sentence from the text into the CLIP encoders, obtaining different features in the temporal dimension. Then, we take the temporal average of the image features and use a text dynamic encoder to obtain a single set of video and text features. We calculate similarity between these features to obtain similarity scores. In this step, we obtain initial multi-dimensional expression label text embeddings and text dynamic encoders. In the second step, we freeze the CLIP encoders and the previously trained text embeddings and text dynamic encoders, while training new video dynamic encoders layer by layer. In the third step, we merge the text and video dynamic encoders to obtain a joint dynamic alignment synchronizer and simultaneously update high-dimensional text embeddings. Thus, we obtain A$^{3}$lign-DFER, where high-dimensional expression label tokens align with dynamic encoding.

\vspace{-0.1cm}
\subsection{Inference}
\label{sec:Inference}

In the inference process, we input MAT and facial expression videos into the frozen CLIP text encoder and image encoder, respectively, on a timestep basis. The features acquired are fed into the JAS along the time dimension, resulting in the extraction of features specific to the video sample and distinct facial expression classes. We compute the similarity between the video features and the features of the expression classes. The class that has the highest prediction probability $cls_{i}$ is the predicted class.


\vspace{-0.1cm}
\section{Experiment}
\label{sec:experiment}

\vspace{-0.1cm}
\subsection{Experiment setup}

\textbf{Datasets.} Our study assesses the effectiveness of A$^{3}$lign-DFER within the context of three real-world DFER datasets: DFEW~\cite{jiang2020dfew}, FERV39k~\cite{wang2022ferv39k}, and MAFW~\cite{liu2022mafw}. We employ DFEW and FERV39k for a 7-class DFER task and utilize MAFW for an 11-class DFER task. Our evaluation sheds light on the performance of our proposed approach when applied to these challenging datasets.

\textbf{Implementation Details.} An A$^{3}$lign-DFER method processes a pair of 16 facial images obtained from consecutive video frames. To enhance computational efficiency, we preprocess all original images by applying the face detection method. This preprocessing generates facial images that are resized to 120 × 120 and then subjected to random cropping to achieve dimensions of 112 × 112 through scale jittering. The proposed framework is implemented on GeForce RTX 3090Ti GPUs with PyTorch toolbox. In the BAP, for the first step, the maximum number of epochs is set to 40, while for the second and third steps, it is set to 30. All models use stochastic gradient descent (SGD) with a Momentum of 0.9, while maintaining a batch size of 64. We use ViT-B/32 and ViT-L/14 of CLIP as the backbone. In our experiments, we configured the hyper-parameters of the MAT as S=16 and T=64. The evaluation metrics utilized include weighted average recall (WAR) and unweighted average recall (UAR).

\vspace{-0.1cm}
\subsection{Comparison with Current Methods}

\begin{table*}[th]
  \caption{Comparison results (\%). The \underline{underscored data} is the second-best result.  \textbf{Our proposed A$^{3}$lign-DFER is the best on three datasets.}}
  \label{tab:methodsota}
  \centering
  \renewcommand{\arraystretch}{1.08}
  \scalebox{0.8}{
      \begin{tabular}{ccccccccc}
        \toprule \toprule 
        \multicolumn{1}{c|}{\multirow{2}{*}{Method}} & \multicolumn{1}{c|}{\multirow{2}{*}{Publication}} & \multicolumn{1}{c|}{\multirow{2}{*}{Backbone}} & \multicolumn{2}{c|}{DFEW}                    & \multicolumn{2}{c|}{FERV39k}  & \multicolumn{2}{c}{MAFW}                    \\  
        \multicolumn{1}{c|}{}                & \multicolumn{1}{c|}{}        & \multicolumn{1}{c|}{}                        & \multicolumn{1}{c}{WAR}   & \multicolumn{1}{c|}{UAR} & \multicolumn{1}{c}{WAR}   & \multicolumn{1}{c|}{UAR}  & \multicolumn{1}{c}{WAR}   & \multicolumn{1}{c}{UAR}   \\ \midrule 
        \multicolumn{1}{c|}{VGG13-LSTM}       & \multicolumn{1}{c|}{/}                           & \multicolumn{1}{c|}{VGG13-LSTM}                                  & \multicolumn{1}{c}{39.05} & \multicolumn{1}{c|}{51.14} & \multicolumn{1}{c}{40.86} & \multicolumn{1}{c|}{29.57} & \multicolumn{1}{c}{/} & \multicolumn{1}{c}{/} \\
        \multicolumn{1}{c|}{C3D~\cite{tran2015learning}}                                  & \multicolumn{1}{c|}{CVPR'15}                                  & \multicolumn{1}{c|}{C3D}                                  & \multicolumn{1}{c}{53.54} & \multicolumn{1}{c|}{42.74} & \multicolumn{1}{c}{ 31.69} & \multicolumn{1}{c|}{22.68} & 42.25 & 31.17 \\
        \multicolumn{1}{c|}{P3D~\cite{qiu2017learning}}                                  & \multicolumn{1}{c|}{ICCV'17}                                  & \multicolumn{1}{c|}{P3D}                                  & \multicolumn{1}{c}{54.47} & \multicolumn{1}{c|}{43.97} & \multicolumn{1}{c}{ 33.39} & \multicolumn{1}{c|}{23.20} & \multicolumn{1}{c}{/} & \multicolumn{1}{c}{/} \\
        \multicolumn{1}{c|}{I3D-RGB~\cite{carreira2017quo}}                                     & \multicolumn{1}{c|}{CVPR'17}                                     & \multicolumn{1}{c|}{Inflated 3D ConvNets}                                     &54.27                     & \multicolumn{1}{c|}{43.40}                     & 38.78                     & \multicolumn{1}{c|}{30.17} & \multicolumn{1}{c}{/} & \multicolumn{1}{c}{/}                     \\
        \multicolumn{1}{c|}{3D ResNet18~\cite{hara2018can}}                                 & \multicolumn{1}{c|}{CVPR'18}                                 & \multicolumn{1}{c|}{ResNet18}                                 & 58.27                     & \multicolumn{1}{c|}{46.52}                     & 37.57                     & \multicolumn{1}{c|}{26.67} & \multicolumn{1}{c}{/} & \multicolumn{1}{c}{/}                    \\
        \multicolumn{1}{c|}{R(2+1)D18~\cite{tran2018closer}}                                 & \multicolumn{1}{c|}{CVPR'18}                                 & \multicolumn{1}{c|}{R(2+1)D}                                 & 53.22                     & \multicolumn{1}{c|}{42.79}                     & 41.28                     & \multicolumn{1}{c|}{31.55} & \multicolumn{1}{c}{/} & \multicolumn{1}{c}{/}                     \\
        \multicolumn{1}{c|}{ResNet18-LSTM~\cite{he2016deep, memory2010long}}                                 &  \multicolumn{1}{c|}{/}                                 & \multicolumn{1}{c|}{ResNet18-LSTM}                                 & 63.85                     & \multicolumn{1}{c|}{51.32}                     & 42.95                     & \multicolumn{1}{c|}{30.92} & 39.38 & 28.08                    \\
        \multicolumn{1}{c|}{ResNet18-ViT~\cite{he2016deep, dosovitskiy2020image}}                                 & \multicolumn{1}{c|}{/}                                 & \multicolumn{1}{c|}{ResNet18-ViT}                                 & 66.56                     & \multicolumn{1}{c|}{55.76}                     & 48.43                     & \multicolumn{1}{c|}{38.35} & 47.72 & 35.80                    \\
        \multicolumn{1}{c|}{EC-STFL~\cite{jiang2020dfew}}                                     & \multicolumn{1}{c|}{MM'20}                                     & \multicolumn{1}{c|}{C3D / P3D / et al.}                                     & 56.51                     & \multicolumn{1}{c|}{45.35}                     &    \multicolumn{1}{c}{/}             &    \multicolumn{1}{c|}{/} & \multicolumn{1}{c}{/} & \multicolumn{1}{c}{/}                      \\
        \multicolumn{1}{c|}{Former-DFER~\cite{zhao2021former}}                                 & \multicolumn{1}{c|}{MM'21}                                 & \multicolumn{1}{c|}{Transformer}                                 & 65.7                      & \multicolumn{1}{c|}{53.69}                     & 46.85                     & \multicolumn{1}{c|}{37.2} & 43.27 &  31.16                    \\
        \multicolumn{1}{c|}{STT~\cite{ma2022spatio}}                                        & \multicolumn{1}{c|}{arXiv'22}                                         & \multicolumn{1}{c|}{ResNet18}                                         & 33.99                     & \multicolumn{1}{c|}{54.58}                     & 48.11                     & \multicolumn{1}{c|}{37.76} & \multicolumn{1}{c}{/} & \multicolumn{1}{c}{/}                     \\
        \multicolumn{1}{c|}{NR-DFERNet~\cite{li2022nr}}                                  & \multicolumn{1}{c|}{arXiv'22}                                  & \multicolumn{1}{c|}{CNN-Transformer}                                  & 68.19                     & \multicolumn{1}{c|}{54.21}                     & 45.97                     & \multicolumn{1}{c|}{33.99} & \multicolumn{1}{c}{/} & \multicolumn{1}{c}{/}                     \\
        \multicolumn{1}{c|}{DPCNet~\cite{wang2022dpcnet}}                                      & \multicolumn{1}{c|}{MM'22}                                      & \multicolumn{1}{c|}{ResNet50 (First 5 Layers)}                                      & 66.32                     & \multicolumn{1}{c|}{57.11}                     &  \multicolumn{1}{c}{/}               &    \multicolumn{1}{c|}{/} & \multicolumn{1}{c}{/} & \multicolumn{1}{c}{/}                      \\
        \multicolumn{1}{c|}{T-ESFL~\cite{liu2022mafw}}                                      & \multicolumn{1}{c|}{MM'22}                                      & \multicolumn{1}{c|}{ResNet-Transformer}                                      & \multicolumn{1}{c}{/}                     & \multicolumn{1}{c|}{/}                     &  \multicolumn{1}{c}{/}               &    \multicolumn{1}{c|}{/} & 48.18 &  33.28                    \\
        \multicolumn{1}{c|}{EST~\cite{liu2023expression}}                                      & \multicolumn{1}{c|}{PR'22}                                      & \multicolumn{1}{c|}{ResNet18}                                      & 65.85                     & \multicolumn{1}{c|}{53.94}                     &  \multicolumn{1}{c}{/}               &    \multicolumn{1}{c|}{/} & \multicolumn{1}{c}{/} & \multicolumn{1}{c}{/}                      \\
        \multicolumn{1}{c|}{Freq-HD~\cite{tao2023freq}}                                      & \multicolumn{1}{c|}{MM'23}                                      &  \multicolumn{1}{c|}{VGG13-LSTM / et al.}                                      &  55.68                     & \multicolumn{1}{c|}{46.85}                     &  \multicolumn{1}{c}{45.26}               &    \multicolumn{1}{c|}{33.07} & \multicolumn{1}{c}{/} & \multicolumn{1}{c}{/}                      \\
        \multicolumn{1}{c|}{LOGO-Former~\cite{ma2023logo}}                                         & \multicolumn{1}{c|}{ICASSP'23}                                         & \multicolumn{1}{c|}{ResNet18}                                         & 66.98                     & \multicolumn{1}{c|}{54.21}                     & 48.13                     & \multicolumn{1}{c|}{38.22} & \multicolumn{1}{c}{/} & \multicolumn{1}{c}{/}                    \\
        \multicolumn{1}{c|}{IAL~\cite{li2023intensity}}                                         & \multicolumn{1}{c|}{AAAI'23}                                         & \multicolumn{1}{c|}{ResNet18}                                         & 69.24                     & \multicolumn{1}{c|}{55.71}                     & 48.54                     & \multicolumn{1}{c|}{35.82} & \multicolumn{1}{c}{/} & \multicolumn{1}{c}{/}                     \\
        \multicolumn{1}{c|}{AEN~\cite{lee2023frame}}                                      & \multicolumn{1}{c|}{CVPRW'23}                                      & \multicolumn{1}{c|}{ResNet18}                                      & 69.37                     & \multicolumn{1}{c|}{56.66}                      & 47.88                     & \multicolumn{1}{c|}{38.18} & \multicolumn{1}{c}{/} & \multicolumn{1}{c}{/}                     \\   
        \multicolumn{1}{c|}{M3DFEL~\cite{wang2023rethinking}}                                      & \multicolumn{1}{c|}{CVPR'23}                                      & \multicolumn{1}{c|}{ResNet18-3D}                                      & 69.25                     & \multicolumn{1}{c|}{56.1}                      & 47.67                     & \multicolumn{1}{c|}{35.94} & \multicolumn{1}{c}{/} & \multicolumn{1}{c}{/}                     \\ 
        \multicolumn{1}{c|}{CLIPER~\cite{li2023cliper}}                                      & \multicolumn{1}{c|}{arXiv'23}                                      & \multicolumn{1}{c|}{CLIP}                                      & 70.84                     & \multicolumn{1}{c|}{57.56}                     & 51.34                     & \multicolumn{1}{c|}{41.23} & \multicolumn{1}{c}{/} & \multicolumn{1}{c}{/}                     \\
        \multicolumn{1}{c|}{DFER-CLIP~\cite{zhao2023prompting}}                                      & \multicolumn{1}{c|}{arXiv'23}                                      &  \multicolumn{1}{c|}{CLIP}                                      & \ul{71.25}                     & \multicolumn{1}{c|}{\ul{59.61}}                     & \ul{51.65}                    & \multicolumn{1}{c|}{\ul{41.27}} & \ul{52.55} & \ul{39.89}                    \\
        \midrule 
        \rowcolor{gray!20} \multicolumn{2}{c|}{\textbf{A$^{3}$lign-DFER (ours)}}                                  & \multicolumn{1}{c|}{CLIP}                                  &    \textbf{74.20}                       &  \multicolumn{1}{c|}{\textbf{64.09}}                         &  \textbf{51.77}                         &  \multicolumn{1}{c|}{\textbf{41.87}} & \textbf{53.24} & \textbf{42.07}        \\    \bottomrule   \bottomrule        
      \end{tabular}
  }
\end{table*}

In the context of our methodology, experiments are conducted on three established DFER datasets, including DFEW~\cite{jiang2020dfew}, FERV39k~\cite{wang2022ferv39k}, and MAFW~\cite{liu2022mafw}. As presented in Table~\ref{tab:methodsota}, the A$^{3}$lign-DFER approach achieves state-of-the-art results on all these datasets. It consistently excels in both the WAR and UAR metrics, demonstrating the efficacy of the method in achieving optimal performance across multiple datasets. For instance, on the DFEW dataset, the approach surpasses the best alternative method DFER-CLIP~\cite{zhao2023prompting}, excluding our own, by 2.95\%$\uparrow$ in WAR and 4.48\%$\uparrow$ in UAR. Specifically, the results of our method outperform all other CLIP-based DFER methods on three datasets.

\begin{table*}[h]
  \caption{Comparison with baseline methods (\%). $\uparrow$ represents a performance improvement compared to CLIP (zero-shot).}
  \label{tab:baseline}
  \centering
  \renewcommand{\arraystretch}{1.08}
  \scalebox{0.85}{
      \begin{tabular}{ccccc}
        \toprule \toprule 
        \multicolumn{1}{c|}{\multirow{2}{*}{Method}}                     & \multicolumn{2}{c|}{FERV39k}  & \multicolumn{2}{c}{MAFW}                    \\  
        \multicolumn{1}{c|}{}                         & \multicolumn{1}{c}{WAR}   & \multicolumn{1}{c|}{UAR}  & \multicolumn{1}{c}{WAR}   & \multicolumn{1}{c}{UAR}   \\ \midrule
        
        \multicolumn{1}{c|}{CLIP (zero-shot)}                                               &  17.09                     & \multicolumn{1}{c|}{20.99} & 19.16 & 18.42                    \\
        \multicolumn{1}{c|}{CoOp}                                          & 42.55(25.46$\uparrow$)                     & \multicolumn{1}{c|}{31.72(10.73$\uparrow$)} & 42.77(23.61$\uparrow$) & 30.79(12.37$\uparrow$)                    \\
        \multicolumn{1}{c|}{CoCoOp}                                        &  44.25(27.16$\uparrow$)                     & \multicolumn{1}{c|}{32.91(11.92$\uparrow$)} &  43.23(24.07$\uparrow$) & 30.81(12.39$\uparrow$)                    \\   \midrule 
        \rowcolor{gray!20} \multicolumn{1}{c|}{\textbf{A$^{3}$lign-DFER (ours)}}                                               &  \textbf{51.77}(34.68$\uparrow$)                         &  \multicolumn{1}{c|}{\textbf{41.87}(20.88$\uparrow$)} & \textbf{53.24}(34.08$\uparrow$) & \textbf{42.07}(23.65$\uparrow$)         \\    \bottomrule \bottomrule       
      \end{tabular}
  }

\end{table*}

\textbf{Effectiveness Beyond CLIP.} It's noteworthy that, The effectiveness of our method stems from the modules we designed, rather than solely from the CLIP pre-trained model. Compared to CLIP (zero-shot) and CLIP-based prompt learning and answer learning methods like CoOp~\cite{zhou2022learning}, CoCoOp~\cite{zhou2022conditional}, the performance shows substantial enhancements, as shown in Table~\ref{tab:baseline}. This is attributed to the method's ability to align video and learnable text in dynamic and affective dimensions, enabling full use of CLIP's prior knowledge and effective adaptation to the evolving demands of DFER downstream tasks. Moreover, our performance surpasses other CLIP-based DFER methods, such as CLIPER~\cite{li2023cliper} and DFER-CLIP~\cite{zhao2023prompting} shown in Table~\ref{tab:methodsota}, while only requiring training on a few modules we proposed, making it more efficient compared to methods like CLIPER~\cite{li2023cliper} that fine-tune CLIP image encoder (4-15 times larger than our modules).



\vspace{-0.1cm}
\subsection{Ablation Study}

In this work, we primarily utilize FERV39k and MAFW for ablation studies, as they currently represent the largest and most challenging datasets for 7-class and 11-class DFER tasks, respectively.

\begin{table*}[th]
  \caption{Ablation study (\%) of different modules in A$^{3}$lign-DFER. $\uparrow$ represents a performance improvement compared to not adding the two modules.}
  \label{tab:moduleablation}
  \centering
  \renewcommand{\arraystretch}{1.08}
  \scalebox{0.85}{
      \begin{tabular}{ll|cccc}
    \toprule \toprule 
    \multicolumn{1}{c}{\multirow{2}{*}{MAT}} & \multicolumn{1}{c|}{\multirow{2}{*}{JAS}} & \multicolumn{2}{c|}{FERV39k} & \multicolumn{2}{c}{MAFW} \\
    \multicolumn{1}{c}{}                      & \multicolumn{1}{c|}{}                      & WAR          & \multicolumn{1}{c|}{UAR} & WAR & UAR       \\ \midrule
    \XSolidBrush                                      & \XSolidBrush                                           &  46.53             &  \multicolumn{1}{c|}{33.24}  & 47.13 & 33.64       \\
    \Checkmark   $_{Average}$                                      & \XSolidBrush                                                 &  48.61(2.08$\uparrow$)            & \multicolumn{1}{c|}{37.33(4.09$\uparrow$)} & 48.52(1.39$\uparrow$) & 36.61(2.97$\uparrow$)          \\
    \XSolidBrush                                         & \Checkmark  $_{Image}$                                              &  48.57(2.04$\uparrow$)            & \multicolumn{1}{c|}{35.29(2.05$\uparrow$)}  & 49.04(1.91$\uparrow$) & 38.72(5.08$\uparrow$)          \\ \midrule
    \rowcolor{gray!20} \Checkmark                                         & \Checkmark                                        &   \textbf{51.77}(5.24$\uparrow$)          & \multicolumn{1}{c|}{\textbf{41.87}(8.63$\uparrow$)} & \textbf{53.24}(6.11$\uparrow$) & \textbf{42.07}(8.43$\uparrow$)            \\    \bottomrule   \bottomrule
    \end{tabular}
  }
\end{table*}

\textbf{Ablation Study of Different Modules.} Ablation experiments on two modules within the A$^{3}$lign-DFER method assess their efficacy, as shown in Table~\ref{tab:moduleablation}. Particularly, for experimental feasibility, when only MAT is available, the average method replaces JAS. When only JAS is present, dynamic processing is exclusively applied to the video. Excluding the MAT and JAS modules for a method relying on a single learnable token and the average of image features results in suboptimal outcomes. However, the introduction of MAT and JAS modules individually substantially enhances the model's performance, convincingly underscoring the effectiveness of affective and dynamic alignment.

\begin{figure}[tb]
  \centering
  \includegraphics[height=5.2cm]{ 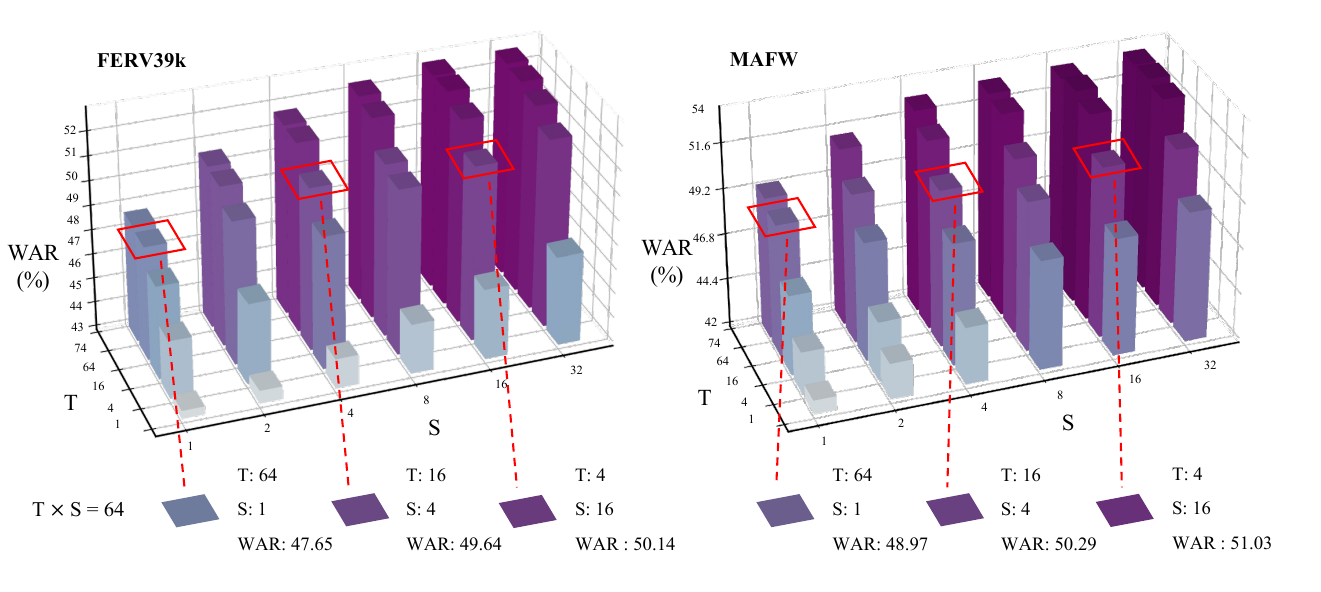}
    \caption{Ablation study of the hyper-parameters of MAT.}
    \label{fig:ablation}
\end{figure}

\textbf{Ablation Study of Hyper-parameters of MAT.} To demonstrate that elevating dimensions for MAT does more than just increase parameters, ablation experiments focusing on the MAT module are conducted, as outlined in Fig.~\ref{fig:ablation}. Expanding dimensions separately for tokens and sentences, the results are summarized in the figure. The horizontal axis represents sentence number $S$ in one class, while the vertical axis represents token dimension $T$ in one sentence. Optimal results in the experiments are achieved with $S = 16$ and $T = 64$. Once the token dimensions reach a certain threshold, it becomes evident that expanding the sentence dimensions, while maintaining the same number of learnable parameters, is more effective. This strategy of dimensional augmentation leads to better outcomes than further increasing the tokens in one dimension, which emphasizes the efficacy of the approach. The MAT method can seamlessly integrate with the subsequent JAS module, effectively aligning dynamic affective information and making the most of joint text and video representations in high-dimensional space. Furthermore, the CLIP pre-trained model imposes constraints on the context length parameter, with a maximum value $T_{\mathrm{max}}$ of 74, removing the $<$SOS$>$, $<$EOS$>$, and the period from the original 77 tokens. The approach allows for the utilization of additional parameters to learn more features from expression videos, effectively reverse-engineering CLIP.

\begin{table*}[th]
  \caption{Ablation study (\%) of the number of transformer layers in JAS, together with the process of BAP. $\uparrow\uparrow$ represents a performance improvement compared to the previous step.}
  \label{tab:JASablation}
  \centering
  \renewcommand{\arraystretch}{1.08}
  \scalebox{0.85}{
      \begin{tabular}{l|cc|cc|cc}
    \toprule \toprule 
    \multicolumn{1}{c|}{\multirow{3}{*}{\makecell{Training \\ Paradigm}}} & \multicolumn{2}{c|}{\multirow{2}{*}{\makecell{Transformer \\ Layers}}}  & \multicolumn{2}{c|}{\multirow{2}{*}{FERV39k}} & \multicolumn{2}{c}{\multirow{2}{*}{MAFW}}\\ 
    \multicolumn{1}{c|}{}                      & \multicolumn{1}{c}{}                      & \multicolumn{1}{c|}{}                            & \multicolumn{2}{c|}{} & \multicolumn{2}{c}{}   \\
    \multicolumn{1}{c|}{}                      & \multicolumn{1}{c}{Image}                      & \multicolumn{1}{c|}{Text}                             & WAR          & \multicolumn{1}{c|}{UAR}   & WAR          & UAR    \\ \midrule
    \multicolumn{1}{c|}{\multirow{2}{*}{\makecell{All at \\ Once}}}                                     &4                                     & 4                                                  &   21.25           &     \multicolumn{1}{c|}{16.09} & 18.14 & 14.27         \\
    \multicolumn{1}{c|}{}                                     &\multicolumn{2}{c|}{\quad 4 (Joint)}                                                              &  20.47            &   \multicolumn{1}{c|}{15.86} & 17.63 & 13.21          \\ \midrule
    \multicolumn{1}{c|}{\multirow{5}{*}{\makecell{\textbf{BAP}}}}                                      
 & Average                                     & 2                                                 &  49.81            &    \multicolumn{1}{c|}{38.03} & 50.52 &  40.10          \\
    \multicolumn{1}{c|}{}                                      &2                                     & 2                                                 &   50.24(0.43$\uparrow\uparrow$ )           &   \multicolumn{1}{c|}{38.69(0.66$\uparrow\uparrow$)} & 51.07(0.55$\uparrow\uparrow$) & 40.72(0.62$\uparrow\uparrow$)           \\
    \multicolumn{1}{c|}{}                                          &4                                         & 2                                                & 50.97(0.73$\uparrow\uparrow$)             &  \multicolumn{1}{c|}{39.03(0.34$\uparrow\uparrow$)} & 52.76(1.69$\uparrow\uparrow$) &  41.69(0.97$\uparrow\uparrow$)           \\
    \multicolumn{1}{c|}{}                                       &4                                      & 4                                              &   51.25(0.28$\uparrow\uparrow$)           &   \multicolumn{1}{c|}{40.38(1.35$\uparrow\uparrow$)} & 52.91(0.15$\uparrow\uparrow$) &  41.85(0.16$\uparrow\uparrow$)         \\  
    \rowcolor{gray!20} \multicolumn{1}{c|}{}                                       &\multicolumn{2}{c|}{\quad 4 (\textbf{JAS})}                                                             &   \textbf{51.77}(0.52$\uparrow\uparrow$)           &  \multicolumn{1}{c|}{\textbf{41.87}(1.49$\uparrow\uparrow$)} & \textbf{53.24}(0.33$\uparrow\uparrow$) & \textbf{42.07}(0.22$\uparrow\uparrow$)   \\  \bottomrule   \bottomrule  
    \end{tabular}
  }
\end{table*}

\textbf{Ablation Study of JAS and BAP.} Regarding the JAS module and BAP approach, relevant ablation experiments evaluate dynamic and bidirectional alignment. As shown in Table~\ref{tab:JASablation}, transformer is employed as the foundational framework for JAS. The methodology begins with an initial phase processing the image using an averaging method and conducting the initial training of the text encoder. After achieving stability in the text dynamic encoder, the image dynamic encoder is progressively trained layer by layer, ultimately reaching the complete structural model. In the final step, a fusion strategy is adopted to fine-tune JAS and the best result is finally gotten. It proves the achievement of dynamic alignment. Direct training of the entire structure all at once does not yield favorable classification results. However, with the implementation of the BAP, classification performance steadily improves, eventually yielding significantly improved results. It proves the effectiveness of bidirectional alignment.

\begin{wraptable}{r}{0.5\textwidth} 
  \caption{Test results under different frame orders (\%). $\downarrow$ indicates a performance decrease compared to the normal order.
  }
  \label{tab:frameorder}
  \centering
  \renewcommand{\arraystretch}{1.3}
  \resizebox{0.5\textwidth}{!}{
  \begin{tabular}{c|cc|cc}
    \toprule
    \toprule
    \multicolumn{1}{c|}{\multirow{2}{*}{\makecell{Test \\ Frame Order}}}  & \multicolumn{2}{c|}{FERV39k} & \multicolumn{2}{c}{MAFW}  \\ 
    & WAR & UAR & WAR & UAR \\ \midrule
    \rowcolor{gray!20} Normal       &  51.77                         &  41.87 & 53.24 & 42.07  \\
    Random & 49.24(2.53$\downarrow$) & 38.65(3.22$\downarrow$) & 49.75(3.49$\downarrow$)& 39.12(2.95$\downarrow$)  \\
    \bottomrule
    \bottomrule
  \end{tabular}
  }
\end{wraptable}

\textbf{Effectiveness of Dynamic Alignment}. As shown in Table~\ref{tab:frameorder}, after training is completed, we shuffle the frame order of the test videos, randomly rearranging them before feeding them into the model for prediction. It was observed that, compared to the normal frame sequence, the prediction accuracy decreased after shuffling. This indirectly indicates that the model has indeed acquired certain dynamic features, considering temporal information as an important reference for classification.

\vspace{-0.1cm}
\subsection{Visualization}

\begin{table*}[th]
  \caption{Visualization of MAT on FERV39k. On the left are the six expressive classes out of the seven facial expression classifications, while on the right are the texts most similar in the embedding space to the tokens learned by MAT.}
  \label{tab:tokenvisualization}
  \centering
  \renewcommand{\arraystretch}{1} 
  \scalebox{0.85}{
      \begin{tabular}{c|llllll}
    \toprule \toprule 
    \multirow{2}{*}{Angry} & imposed & \emph{ruined} & backtalked & twits & bombings & cracked \\
                           & blasted & \emph{frenzy} & tense & kidnap & fuel & dramatic \\
    \midrule
    \multirow{2}{*}{Disgust} & friggin & \textbf{disgusting} & stain & \emph{disgraceful} & prejudice & filthy \\
                             & \emph{damned} & \emph{gruesome} & \emph{daft} & faults & ills & messing \\
    \midrule
    \multirow{2}{*}{Fear} & fled & sos & \emph{pressing} & fights & slaughtered & \textbf{fear} \\
                          & misery & vampire & struggling & infected & intimidating & fright \\
    \midrule
    \multirow{2}{*}{Happy} & feat & nursery & hobby & \emph{emotionally} & \textbf{happy} & smile \\
                           & winner & wedding & kinder & hopefully & praise & champ \\
    \midrule
    \multirow{2}{*}{Sad} & loser & complain & denying & \emph{hard} & \emph{ethereal} & asocial \\
                         & sleepless & \emph{gloom} & patientrooms & sacrificing & mourn & fever \\
    \midrule
    \multirow{2}{*}{Surprise} & shout & \emph{excite} & giveaway & \textbf{surprise} & campground & miraculous \\
                              & \emph{unpredictable} & comet & severance & romantic & innovations & imposing \\
    \bottomrule \bottomrule 
    \end{tabular}
  }
\end{table*}

\textbf{Visualization of Tokens in MAT.} Table~\ref{tab:tokenvisualization} shows distinct features learned by MAT. We identified vocabulary words highly similar to these embeddings, focusing on expressions and excluding irrelevant terms like names of people and places. MAT successfully learned words related to six out of seven expression classes, certain embeddings exactly matching abstract expression descriptions (\textbf{bolded} in the table). Affection-related terms are \emph{italicized}, highlighting MAT's precision. The A$^{3}$lign-DFER method's effective alignment of abstract affections with expression videos is visually represented. These rich and multidimensional embeddings help improve the performance of DFER.

\begin{figure}[th]
  \centering
  \includegraphics[height=9cm]{ 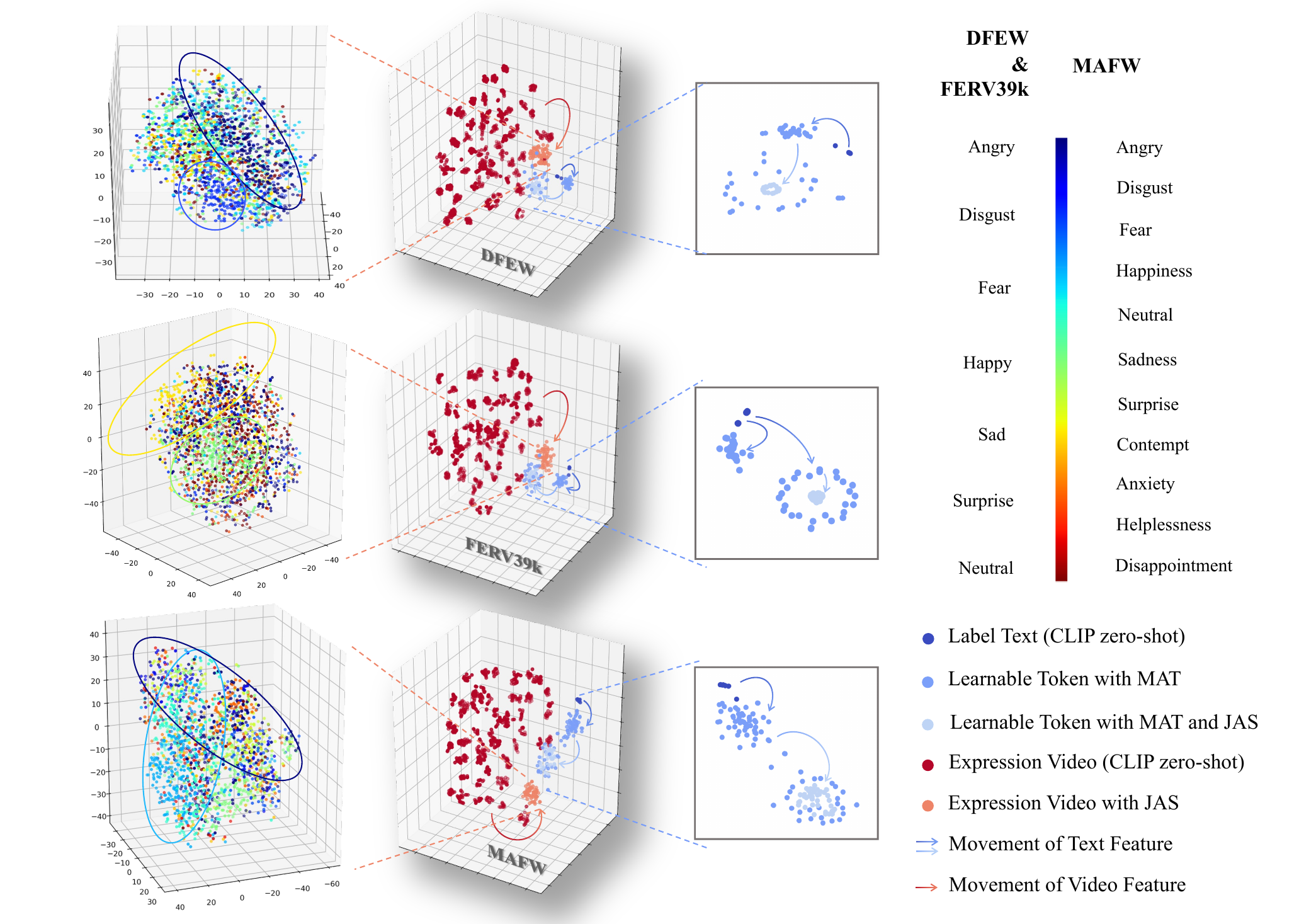}

   \caption{The t-SNE visualizations on the Test Data of the DFEW, FERV39k and MAFW datasets. In each visualization figure, the left image displays the t-SNE results of the A$^{3}$lign-DFER process, the upper right image details the t-SNE results for video sample classes, and the lower right image details the t-SNE results of the text flow.}
   \label{fig:tsne}
\end{figure}

\textbf{t-SNE Visualization of High-level Features.} Fig.~\ref{fig:tsne} presents t-SNE reduced features from three datasets, showing the A$^{3}$lign-DFER process. Initially distant in CLIP's feature space, image and text features converge through MAT processing, aligning with classification goals. The lower right image details text data flow, while the upper right shows post-processed video sample features, demonstrating intra-class aggregation and inter-class separation. This approach, focusing on text-image similarity in high-dimensional space, sometimes results in overlapping features, yet maintains strong class aggregation, marked by circles.


\vspace{-0.1cm}
\section{Conclusion and Discussion}
\label{sec:conlusion}

This paper introduces A$^{3}$lign-DFER as a notable breakthrough in DFER, effectively overcoming alignment challenges. Utilizing innovative components like MAT, JAS, and BAP, our method comprehensively aligns dynamic affective responses, enhancing CLIP's utility for DFER and setting new affective computing benchmarks. A$^{3}$lign-DFER's effectiveness is confirmed by state-of-the-art results on datasets like DFEW, FERV39k, and MAFW, and further supported by detailed ablation studies and visualization analyses. 

These accomplishments underscore A$^{3}$lign-DFER's role as a pioneering solution, setting a new standard in the realm of DFER. Going forward, future efforts will be dedicated to enhancing its functionality. A primary objective is to overcome the limitation of requiring training for each expression class, which currently restricts its adaptability to emerging, subtler expressions. We plan to investigate strategies like broadening affective labels to maximize CLIP's inherent capabilities and create a zero-shot DFER model. Additionally, conducting comprehensive scientific research on full alignment is also a crucial direction.

%
%
\bibliographystyle{splncs04}
\bibliography{main}
\end{document}